\documentclass[lettersize,journal]{IEEEtran}
\usepackage{amsmath,amsfonts}
\usepackage{algorithm}
\usepackage{algorithmicx}
\usepackage{algpseudocode}
\usepackage{array}
\usepackage[caption=false,font=normalsize,labelfont=sf,textfont=sf]{subfig}
\usepackage{textcomp}
\usepackage{stfloats}
\usepackage{url}
\usepackage{verbatim}
\usepackage{graphicx}
\usepackage{cite}
\usepackage{booktabs}
\usepackage{multirow}
\usepackage{setspace}
\usepackage{amsmath}
\usepackage{amsthm}
\usepackage{wrapfig}
\usepackage{subfig,color} 
\usepackage{xcolor}

\allowdisplaybreaks

\makeatletter
\newcommand*\bigcdot{\mathpalette\bigcdot@{.5}}
\newcommand*\bigcdot@[2]{\mathbin{\vcenter{\hbox{\scalebox{#2}{$\m@th#1\bullet$}}}}}{\large }
\makeatother

\begin{document}

\title{CORB-Planner: Corridor as Observations for RL Planning in High-Speed Flight}
\author{Yechen Zhang, Bin Gao, Gang Wang,  Jian~Sun, and Zhuo Li
		\thanks{The work was supported in part by the National Natural Science Foundation of China under Grants U23B2059, 62173034, 62495090, 62495095, and 62088101.
  }
    
		\thanks{Yechen Zhang, Bin Gao,  Gang Wang, Jian Sun, and Zhuo Li are with the State Key Laboratory of Autonomous Intelligent Unmanned Systems, School of Automation, Beijing Institute of Technology, Beijing 100081, China. (e-mail: yczhang@bit.edu.cn; gaobin@bit.edu.cn; gangwang@bit.edu.cn; sunjian@bit.edu.cn;   zhuoli@bit.edu.cn).
}
}


\maketitle

\allowdisplaybreaks

\begin{abstract}

Reinforcement learning (RL) has shown promise in a large number of robotic control tasks. Nevertheless, its deployment on unmanned aerial vehicles (UAVs) remains challenging, mainly because of reliance on accurate dynamic models and platform-specific sensing, which hinders cross-platform transfer. This paper presents the CORB-Planner (Corridor-as-Observations for RL B-spline planner), a real-time, RL-based trajectory planning framework for high-speed autonomous UAV flight across heterogeneous platforms. The key idea is to combine B-spline trajectory generation—with the RL policy producing successive control points—with a compact safe flight corridor (SFC) representation obtained via heuristic search. The SFC abstracts obstacle information in a low-dimensional form, mitigating overfitting to platform-specific details and reducing sensitivity to model inaccuracies. To narrow the sim-to-real gap, we adopt an easy-to-hard progressive training pipeline in simulation. A value-based soft decomposed-critic Q (SDCQ) algorithm is used to learn effective policies within approximately ten minutes of training. Benchmarks in simulation and real-world tests demonstrate real-time planning on lightweight onboard hardware and support maximum flight speeds up to $8.2$m/s in dense, cluttered environments without external positioning. Compatibility with various UAV configurations (quadrotors, hexarotors) and modest onboard compute underlines the generality and robustness of CORB-Planner for practical deployment.

\end{abstract}

\allowdisplaybreaks

\section{Introduction}

Reinforcement learning (RL) has emerged as a powerful paradigm for robotic control and planning, yielding significant advancements in applications ranging from bipedal locomotion \cite{haarnoja2024learning} to quadrupedal motion \cite{lee2020learning}. With appropriate safety measures, RL agents can be trained directly on physical platforms, enabling efficient policy learning for complex tasks. Extending RL to UAVs, however, presents additional challenges. Training directly on physical UAVs incurs high cost and safety risk, and sim-to-real transfer is impeded by discrepancies in dynamics and environment. Previous approaches have addressed these issues by leveraging high-fidelity dynamic models \cite{kaufmann2023champion, Zhang2024Back,zhou2023efficient} or by constraining RL to high-level trajectory planning \cite{loquercio2021learning}. Nevertheless, such methods often depend on platform-specific sensors (e.g., depth cameras or LiDAR) and detailed dynamic parameters, limiting cross-platform adaptability.

In this paper, we introduce CORB-Planner, a simple and cross-platform B-spline trajectory planning framework that integrates RL for efficient, platform-agnostic UAV trajectory generation. Recognizing that state-of-the-art control methods achieve accurate trajectory tracking \cite{10778610,9555602,gkx2022ral}, we decouple trajectory planning from navigation and flight control. This decoupling reduces the computational demands on board and enhances the generalizability of RL-based planners. Our framework employs grid maps and odometry data generated with various sensor configurations (e.g., Fast-LIO2 \cite{10757429}, Point-LIO \cite{he2023point} and VINS-Fusion \cite{qin2019general}). A reference polyline is initially computed using the 3-dimensional $A^*$ search algorithm, around which an SFC is constructed \cite{liu2017planning}.  SFC encodes obstacle information using fewer than 100 features, thereby mitigating overfitting to environment-specific details. The RL agent leverages the SFC along with an initial set of B-spline control points to iteratively generate refined control points in a first-in, first-out manner. This process yields smooth trajectories comprising $10$ to $15$ control points, which are subsequently tracked by the UAV’s flight controller.

The training of the CORB-Planner agents is accomplished in a simplified environment devoid of physical simulations. During the training phase, we employed the SDCQ algorithm, which we previously proposed in \cite{zhang2023soft}. This algorithm finely discretizes the three-dimensional action space and utilizes a deep Q-network to approximate discrete Q-values. Concurrently, a continuous critic network is deployed for target evaluation, facilitating efficient training through supervised learning between continuous and discrete Q-functions. To enhance performance in narrow and cluttered environments, we designed a training environment with progressive difficulty and augmented exploration by sampling multiple exploratory trajectories while utilizing non-exploratory trajectories for flight control. This approach accelerates convergence and mitigates overfitting. After a training duration of ten minutes, the CORB-Planner training system yields effective planning agents applicable to a broad spectrum of UAV platforms.

\begin{figure*}[t]\label{fig1}
	\centering
	\includegraphics[width=7in]{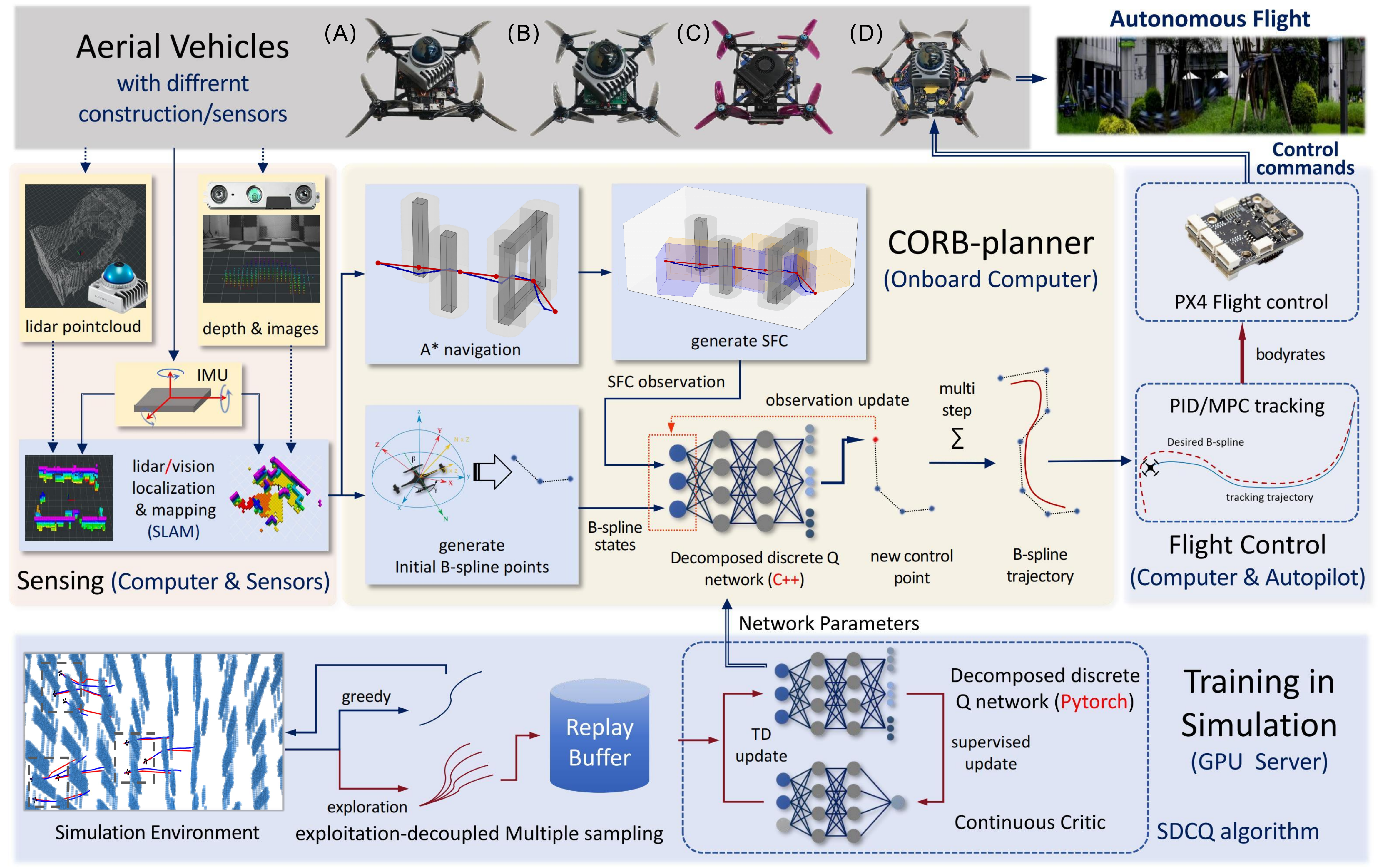}
    	\caption{System architecture of CORB-Planner. The preprocessing module constructs the SFC and B-spline initialization from onboard odometry and grid maps. The training module uses an easy-to-hard curriculum and the SDCQ algorithm to produce the planner’s network parameters. The online planner then sequentially generates B-spline control points within the SFC, and the flight control system tracks the resulting trajectory in real time.} 
\label{rlsystem}
\end{figure*}

Extensive evaluation confirms the proposed CORB‑Planner’s effectiveness: in simulation, SDCQ‑trained agents converge within ten minutes and outperform state‑of‑the‑art optimization‑based planners across diverse scenarios. In real‑world flights—on LiDAR‑equipped quadrotors (Livox MID360), vision‑based quadrotors (Realsense D435i), and hexarotors—CORB-Planner navigates dense clutter at speeds up to $8.2$m/s, running entirely on a lightweight $65$mm$\times30$mm board powered by a quad‑core Cortex‑A53. Fig. \ref{rlsystem} illustrates the end‑to‑end architecture, encompassing preprocessing, training, and deployment.

The main contributions of this paper are summarized as follows.

\begin{itemize}
\item Lightweight, cross‑platform planning: CORB‑Planner combines low‑dimensional SFC with RL‑driven B‑spline control‑point generation, enabling real‑time trajectory planning on diverse UAVs with minimal onboard compute. 
\item Ten‑minute, curriculum‑based training: Our easy‑to‑hard simulation pipeline, coupled with the SDCQ algorithm and an exploitation‑decoupled exploration strategy, produces robust planners in under ten minutes without physics simulation.
\end{itemize}

\section{Related Work} 
\subsection{Reinforcement Learning for Aerial Systems} 

Recent advancements in RL have shown applicability to UAV control, especially for low-level flight dynamics. The Swift system \cite{kaufmann2023champion} uses proximal policy optimization (PPO) to achieve superhuman drone racing performance with visual-inertial odometry (VIO) inputs, reaching speeds of 13.11 m/s, but its end-to-end design limits generalization beyond fixed racing tracks. Differentiable simulation approaches \cite{Zhang2024Back} combine deep learning with first-principles physics for obstacle avoidance, yet often rely on specific sensor suites (e.g., depth cameras), reducing portability. Policy-guided trajectory planning methods \cite{loquercio2021learning} decouple high-level planning from low-level control to generate smooth trajectories, but typically assume depth-based sensing. CORB-Planner builds on these advances by merging RL-based planning with B-spline trajectory generation, abstracting environment details via compact SFC representations to achieve platform-agnostic, real-time planning.

\subsection{B-spline Trajectory Planning} 

B-splines are widely used in trajectory planning for their smoothness and local control. These curves are parameterized by a series of control
points, where higher-order B-splines achieve greater smoothness.  Each control point influences only a local segment, and the trajectory lies within the convex hull of its control points. A uniform third-order B-spline is expressed as
\begin{align}\label{bsplinedescribe} p(\tau) &= \sum_{t=0}^{n}p_t B_{t, 3}(\tau), \quad \tau \in [\tau_{k-1}, \tau_{n+1}], \quad \text{where}~\nonumber\\
B_{t, k}(\tau) &= \frac{\tau - \tau_{t}}{\tau_{t+k-1} - \tau_{t}} B_{t, k-1}(\tau) +
\frac{\tau_{t+k} - \tau}{\tau_{t+k} - \tau_{t+1}} B_{t+1, k-1}(\tau) \notag\\  B_{t, 0}(\tau) &= 
\begin{cases} 1, & \tau_t \leq \tau < \tau_{t+1}\\  
 0,& \text{otherwise}
 \end{cases} \notag \\
\tau_t &= t\Delta\tau
\end{align}
where 
$p$ denotes the position trajectory, $\tau$ represents time, $k$ specifies the spline order, $p_t$  are the control points, $B_{t,k}(\tau)$ are the B-spline basis functions, and $t$ indexes the control points in the sequence. In the CORB-Planner, a new control point is generated at each RL decision step, with 
$t$ corresponding to the RL decision index and $\Delta \tau$ representing the fixed interval between knots.

Both CORB-Planner and the widely adopted EGO-planner \cite{zhou2020ego} utilize third-order B-splines to model vehicle trajectories. These B-splines guarantee continuous acceleration, ensuring trajectories are feasible for flight control. The derivatives of B-splines—lower-order B-splines themselves—enable computation of velocity, acceleration, and jerk control points directly from the position trajectory:
\begin{equation}\label{controlpoints} v_t = \frac{p_{t+1} - p_t}{\Delta \tau}, \quad a_t = \frac{v_{t+1} - v_t}{\Delta \tau}, \quad j_t = \frac{a_{t+1} - a_t}{\Delta \tau},
\end{equation}
where $v_t$, $a_t$, and $j_t$ denote velocity, acceleration, and jerk control points, respectively.
EGO-planner employs heuristic search for obstacle avoidance and optimizes trajectories using a penalty function to evaluate deviations between collision-free guidance paths and collision-prone trajectories. By limiting its focus to local grid map sections relevant to potential collisions, EGO-planner avoids constructing a global Euclidean signed distance field, enabling real-time operation.

\subsection{RL Algorithms for Aerial Systems}
Effective UAV trajectory planning involves controlling continuous quantities—body rates, linear velocities, and B‑spline control points—within tight dynamic constraints. Most state‑of‑the‑art continuous‑action RL methods follow an actor‑critic paradigm. On‑policy approaches like PPO \cite{schulman2017proximal} offer stable updates but can be sample‑inefficient. Off‑policy variants—DDPG \cite{lillicrap2015continuous}, TD3 \cite{fujimoto2018addressing}, and SAC \cite{haarnoja2018soft1,haarnoja2018soft2}—improve efficiency and robustness, yet still struggle on resource‑constrained platforms.
More recent approaches leverage world‑model architectures—for example, Dreamer‑V3 \cite{hafner2025mastering} and Storm \cite{zhang2023storm}—to learn latent dynamics representations that drive policy learning. While these methods achieve impressive generalization and sample efficiency, the additional computational overhead of training and rolling out learned models makes real‑time onboard deployment challenging. To navigate this trade‑off, CORB‑Planner employs our SDCQ  algorithm \cite{zhang2023soft}, which discretizes each action dimension independently but uses a continuous critic for target evaluation. This hybrid design retains the stability and expressiveness of continuous‑action RL while maintaining low computational and memory footprints, making it well suited to resource‑constrained UAV platforms.

\begin{figure*}[t]
	\centering
	\includegraphics[width=6.8in, height=1.7in]{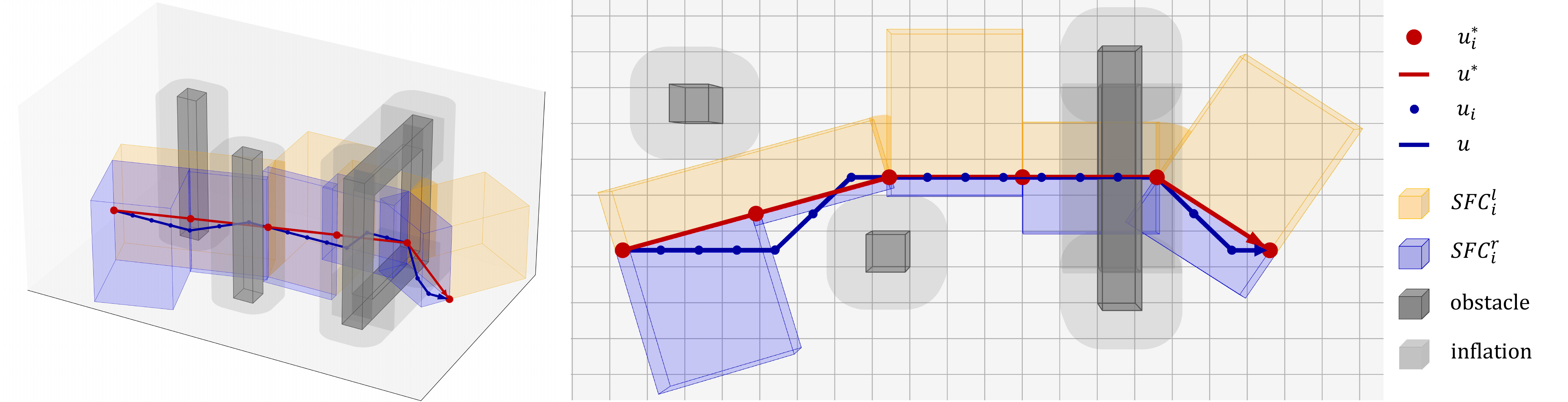}
	\caption{An example of $A^*$ path $u$, reference polyline $u^*$ and the SFC defined in CORB-Planner. Top view of the example is on the right side of the figure. $SFC_i^l$ on the left side of the polyline is represented by the orange cubic, and $SFC_i^r$ on the right side of the polyline by the blue cubic.}
	\label{sfc}
\end{figure*}

\section{RL Formulation of CORB-Planner} 
The CORB-Planner leverages RL to provide a computationally efficient alternative to optimization-based approaches for B-spline trajectory planning. At each decision step, the RL policy generates a control point, thereby reducing the computational overhead associated with trajectory optimization. However, high-dimensional observations such as 3D grid maps pose challenges for RL agents, as neural networks can overfit to extraneous details, adversely affecting generalization.

In optimization-based path planning methods, safe flight corridors are employed to simplify the complexity of obstacle representation\cite{ren2022bubble}, and they also demonstrate potential for application in RL-based trajectory planning approaches. To mitigate the overfitting problem, CORB-Planner employs simplified representations of obstacle environments through the use of SFCs. The RL formulation is structured as a Markov decision process (MDP) defined by the tuple 
$\{\mathcal{S}, \mathcal{A}, \mathcal{P}, r\}$, where $\mathcal{S}: {s_t\in R^o}$ is the observation space, $\mathcal{A}: {\alpha_t\in R^m}$ is the action space, $\mathcal{P}:=\mathcal{P}(s_{t+1}|s_t, \alpha_t)$ is the transition probability, and $r_t$ is the reward function.

The CORB-Planner prioritizes the design of the observation space $\mathcal{S}$, action space $\mathcal{A}$, and reward function $r$. The transition dynamics $\mathcal{P}$ are governed by the interaction of $\mathcal{S}$, $\mathcal{A}$, and the third-order B-spline trajectory model.

\subsection{SFC as Observations for RL}

In CORB-Planner, the observation space $\mathcal{S}$ integrates the drone's motion state, obstacles represented as SFCs, and the planning time $\tau$. The observation at step $t$ is expressed as follows, which can present the obstacle environment and the drone dynamics precisely:
\begin{equation*}
	s_t := [p_{t-2},p_{t-1},p_{t}, SFC, \tau ]. 
\end{equation*}
At knots where  $\tau = \tau_t = t\Delta \tau$, the trajectory state depends solely on three control points, reducing dimensionality and simplifying computation, rather than four points with $\tau\neq t\Delta \tau$. During the initial planning phase, the first three control points of the B-spline are derived from the initial position, velocity, and acceleration:
\begin{equation}
	\frac{p_0}{6} + \frac{2p_1}{3} + \frac{p_2}{6} = p(\tau_2), \ 
	\frac{v_0}{2} + \frac{v_1}{2} = v(\tau_2), \ 
	a_0 = a(\tau_2)
    \label{initialpoints}
\end{equation} 
where the velocity and acceleration control points $v_0, v_1, a_0$ are recursively derived from position control points using \eqref{controlpoints}. Position $p(\tau_2)$, velocity $v(\tau_2)$, and acceleration $a(\tau_2)$ are obtained from the odometry, ensuring a unique solution for uniform B-splines.

 SFC is a crucial component of the state representation in CORB-Planner.
While the planner leverages 3D grid maps, directly incorporating these maps into the state will lead to overfitting and compromise generalization capabilities. To mitigate this issue, we utilize the 3D $A^*$ search algorithm to compute a reference path within the grid map. Based on this path, a simplified SFC is constructed. An example of the SFC in a 3D context is illustrated in Fig. \ref{sfc}.

The reference path generated by $A^*$ is denoted as $u = <u_0 \rightarrow u_1 \rightarrow \cdots \rightarrow u_k>$, where each control point $u_i = [u_{i,x} \ u_{i,y}, u_{i, z}]^\top$ represents a 3D coordinate. Since the path generated by A* may contain unnecessary detours, we select a subset of points $u^*$ from  $u$ to form a polyline with the minimum number of segments possible. The selection process of $u^*$ is demonstrated in Alg. \ref{alg1}.

\begin{algorithm}[h]
	\caption{Get polyline $u^*$ from A* path $u$}
	\label{alg1}
	\begin{algorithmic}[1] 
		\Require grid map, $u = <u_0 \rightarrow u_1 \rightarrow \cdots \rightarrow u_k>$
		\State Initialize $u^* = <u_0>$, local start point $u^*_{start} = u_0$
		\For {$i=2$ {to} $k$}
		\If{segment $<u^*_{start}\rightarrow u_i>$ collided}
		\State Add $u_{i- 1}$ to $u^*$
		\State $u^*_{start} = u_{i - 1}$
		\EndIf
		\State Add $u_{k}$ to $u^*$
		\EndFor
	\end{algorithmic}
\end{algorithm}
As depicted in Fig. \ref{sfc}, $u^*=<u^*_0\rightarrow u^*_1\rightarrow ... \rightarrow u^*_k>$  (depicted as the red polyline) circumvents the detours inherent in $u$ (depicted as the blue polyline). For segments in $u^*$ that exceed a certain length threshold, we will partition these segments into multiple subsections to enhance the precision of the SFC. The path segments in $u^*$ is definced as $L_i = <u^*_{i-1} \rightarrow u^*_i>$, $i \in [1, k]$.


The SFC is composed of sub-corridors $SFC_i$ constructed around each path segment $L_i$. Each sub-corridor is partitioned into left ($SFC_i^l$) and right ($SFC_i^r$) sections based on proximity to the nearest obstacle. To clarify, For each $L_i$ in $u^*$ we defines its corresponding orthogonal unit vector:
\begin{equation}
	n_i = [u^*_{i, y-1} \!-\! u_{i, y} \ \  u_{i, x} \!-\! u_{i, x-1}  \  \  0]^\top / \ |L_i|_2.
\end{equation} 
The SFC of CORB-Planner is defined as:
\begin{align}
SFC = \overset{k}{\underset{i=1}{\cup}}SFC_i = \overset{k}{\underset{i=1}{\cup}}\{SFC^l_i \cup SFC^r_i\}\label{globalsfc}
\end{align}
\begin{align}
 SFC_i^l = \{p \mid (\text{dist}_{xy}(p, L_i) < \Delta_i^l),  \ & ((p - u_i)^\top n_i > 0),\notag\\  
 & (z_{inf} < p_z < z_{sup}) \}  \notag\\
 SFC_i^r = \{p \mid (\text{dist}_{xy}(p, L_i) < \Delta_i^r),  \ & ((p - u_i)^\top n_i < 0),\notag\\
 & (z_{inf} < p_z < z_{sup}) \}.\label{sfcg2}
\end{align}
Here, $p(x,y,z)$  represents a point in 3D space, $\text{dist}_{xy}(p, L_i)$  denotes the Euclidean distance between the 2D projection of $p$ and the line segment  $L_i$, and $p_z$ is the $z$-coordinate of $p$, $\Delta_i^l$ and $\Delta_i^r$ represent the minimum obstacle distances on the left and right sides of $L_i$, within the $z$-axis range defined by $z_{inf} < p_z < z_{sup}$. Given that the values of $\Delta_i^l$ and $\Delta_i^r$ vary across different z-axis ranges, we conduct sampling for multiple $SFC_i$ instances over varying z-axis ranges and select the optimal one that maximizes $(z_{sup} - z_{inf}) \dot (\Delta_i^l + \Delta_i^r)$. Each sub-corridor $SFC_i$ described by \ref{sfcg2} can be characterized using 8 dimensions in the state space $\mathcal{S}$, which include the x-axis and y-axis coordinates of $u^*_{i-1}$ and $u^*_i$, along with $\Delta_i^l, \Delta_i^r, z_{sup}, z_{inf}$. Since the coordinate of $u^*_i$ are shared between $SFC_i$ and $SFC_{i + 1}$, a total of $6k-4$ features are required to fully describe the global SFC with $k$ points and $k-1$ sub-corridors. 

Further, including the entire SFC in the state space may lead to overfitting, as sub-corridors distant from the current position have minimal impact on the immediate decision-making process. The state space in CORB-Planner considers a subset of the SFC, including $N$ sub-corridors directly ahead of the current planning position $p(\tau_t)$ (i.e. if in step $t$, $p(\tau_t)$ locates in $SFC_i$,the corresponding state space including $SFC_i$ to $SFC_{i+N}$). In practice, we set $N=9$, which covers $9$ sub-corridors and $10$ points in $u^*$. This results in a $56$-dimensional observation space. When incorporating time information $\tau_t$ and B-spline points $p_{t-2}, p_{t_1}, p_t$, the total state dimensionality is $66$, ensuring computational efficiency and generalization.

\subsection{Action Space Transformation for Uniformity in Directions}


In CORB-Planner, the action space $\mathcal{A}$ is 3D, corresponding to an 3D acceleration control point, which can be transferred to position control point. Each dimension of $\mathcal{A}$ is constrained by the maximum acceleration $a_{max}$, resulting in an action space where the values for each axis are bounded within the interval $[-a_{max}, a_{max}]$. Notably, the maximum allowable acceleration in the $z$-axis is set to be less than $9.8$ m/s$^2$. This configuration forms a cubic action space. However, such a naive design implies non-uniform maximum accelerations across different directions. For example, the maximum acceleration along the $45^\circ$ diagonal is $\sqrt{2}$ times greater than along the $x$-axis.

To give out a uniform dynamic constraint across different directions, the CORB-Planner employs a transformation that maps the cubic action space into a cylindrical shape, thereby ensuring uniform maximum acceleration across all horizontal directions. The acceleration mapping is described as follows 
\begin{equation}
\label{actionbound} 	
\alpha_x \leftarrow \frac{\alpha_x\max(\alpha_x, \alpha_y)}{ \sqrt{\alpha_x^2 + \alpha_y^2} + \epsilon}; \  \  \alpha_y \leftarrow \frac{\alpha_y\max(\alpha_x, \alpha_y)}{ \sqrt{\alpha_x^2 + \alpha_y^2} + \epsilon} 
\end{equation} 
where $\alpha_t := [\alpha_x \ \alpha_y \ \alpha_z]^\top\in[-1, 1]^3$ is the action output at planning step $t$. The acceleration control point $a_t = \alpha_t a_{max}$ is scaled by the maximum acceleration $a_{max}$. Similarly, velocity constraints are designed to ensure uniformity in maximum velocity across all directions. Let $v_t := [v_x \ v_y \ v_z]^\top$ represent the velocity control point at step $t$, with $v_{max}$ as the maximum velocity. The velocity control under uniform constraints is represented as
\begin{subequations}\label{speedbound}
	\begin{equation}
	v_{t+1} = v_t + \alpha_t \times a_{max} \times  \Delta t
	\end{equation}
	\begin{equation}
	v_x \!\leftarrow\! \frac{v_x v_{max}}{\max(v_{max},\! \sqrt{v_x^2  \!+\! v_y^2})};  \ 
	v_y \!\leftarrow\! \frac{v_y v_{max}}{\max(v_{max},\! \sqrt{v_x^2 \!+\! v_y^2})}.  
	\end{equation}
	
\end{subequations}

Once $v_{t+1}$ is generated, the next position control point $p_{t+1}$ is determined using $v_{t+2}$ and $p_t$. The new position $p_{t+1}$ replaces $p_{t-2}$ in $s_t$ in a first-in-first-out manner. Meanwhile, the local SFC in $s_t$ is updated according to $p(\tau_{t+1})$, and the time parameter $\tau = \tau_t$ is updated to $\tau_{t+1}$, advancing the state from $s_t$ to $s_{t+1}$.

\subsection{Reward Function}

In RL, the reward function guides the agent's trajectory toward desired behaviors, such as safety and smoothness. In the context of UAV flight planning, the reward function must reflect multiple factors, including obstacle avoidance, trajectory smoothness, and flight velocity. To satisfy these objectives, we design a reward function consisting of three key components: an obstacle avoidance penalty $r_p(t)$, a SFC-follow reward $r_f(t)$, and a success reward $r_s(t)$. The smoothness constraints are incorporated into the velocity reward $r_f(t)$. The total reward at time $t$ is a weighted sum of these components, expressed as
\begin{equation} 
r_t = k_p r_p(t) + k_f r_f(t) + k_s r_s(t) 
\end{equation}
where $k_p$, $k_v$, and $k_s$ are the weighting coefficients that balance the contributions of each component. 

The terms $r_p(t)$, $r_f(t)$, and $r_s(t)$ are established through a comparative analysis of the B-spline trajectory against the reference polyline $u^*$ and the SFC. Given that the vehicle is intended to adhere to the SFC, the forward progression of the B-spline can be quantified by pinpointing the specific sub-corridor $SFC_i$ encompassing the B-spline knot position $p(\tau_t)$. For example, if $SFC_m$  denotes the sub-corridor containing $p(\tau_t)$ and $SFC_n$ contains $p(\tau_{t+1})$,  this signifies that the B-spline has traversed $n-m$ sub-corridors from step $t$ to $t+1$, thereby indicating its advancement along the designated path.  

The SFC-follow reward $r_f(t)$ is defined as the cumulative length of polyline segments  $\Vert{L_i}\Vert_2$ that the B-spline traverses within their corresponding sub-corridors during step $t$.  Specifically, if the B-spline knot position $p(\tau_t)$ is situated simultaneously within  $SFC_i$ and $SFC_{i+1}$, it is assigned to $SFC_{i+1}$. If $p(\tau_{t+1})$ falls outside the SFC, the trajectory planning process will be terminated.
\begin{equation} 
r_f(t) = \sum_{i=m}^{n-1}\Vert{L_i}\Vert_2,  \  \text{where} \  p(\tau_t) \in SFC_m, \ p(\tau_{t+1}) \in SFC_n 
\end{equation}

The action space of the CORB-Planner does not incorporate jerk constraints directly. Consequently, soft jerk constraints are integrated into the reward function to ensure trajectory smoothness. To simplify the complexity associated with balancing the components of $r_t$,  a jerk-related discount is introduced on $r_f(t)$, as opposed to incorporating an additional jerk penalty term.  If the jerk $j(\tau_t)$ exceeds half of the maximum allowable jerk $j_{max}$, the SFC-followed reward  is subject to reduction. The application of this discount is detailed as follows
\begin{equation}
	r_f(t)\leftarrow \left\{ \begin{array}{ll} 
r_f(t), &j(\tau_t) \leq \frac{j_{max}}{2} \\
r_f(t)  \frac{2  (j_{max} - j(\tau_t)) }{ j_{max}},  &\frac{j_{max}}{2} < j(\tau_t) \leq j_{max} \\
0,&j(\tau_t) > j_{max}.
\end{array}\right.
\end{equation}

The obstacle avoidance penalty $r_p(t)$ is based on the vehicle’s adherence to the 3D SFC. A negative reward, along with an early termination signal, is issued if the B-spline exceeds the boundaries of the SFC. To detect potential violations between B-spline knots $\tau_{t-1}$ and $\tau_t$, ten points are uniformly sampled along the trajectory to ensure the entire segment remains within the SFC
\begin{align}
	r_p^k(t)& =  \left\{ \begin{array}{ll} 
		0, &p(\tau_{t-1} + k\Delta\tau / 10) \notin  SFC \\
		 1, &p(\tau_{t-1} + k\Delta\tau / 10) \in  SFC
	\end{array}\right. \\
		r_p(t)& = \mathop{\prod^{10}_{k = 1}}r_p^k(t) - 1.
\end{align}

The success reward $r_s(t)$ is given when the B-spline trajectory reaches the last sub-corridor:
\begin{equation}
	r_s(t) = \left\{ \begin{array}{ll}
		1, &\ p(\tau_{t+1}) \in SFC_k \\
		0, &\text{otherwise.}
	\end{array}\right.
\end{equation}

Nevertheless, the trajectory planning process is designed to terminate early if the B-spline exceeds the SFC, indicated by $r_p(t) = -1$, or if it violates the maximum jerk constraint (with velocity and acceleration limited by the action space). We define $D_t$ as the termination indicator at time $t$. Since the reward function is used solely for training purposes, $D_t$ can be employed to evaluate the feasibility of B-splines in real-time applications, reducing the need to compute $r_t$ during execution. The termination indicator $D_t$ is defined as follows
\begin{equation}
		D_t =  \left\{ \begin{array}{ll} 
		1, &r_p(t) = -1 \\[0.05cm]
		1, &j(\tau_t) >j_{max} \\[0.05cm]
		0, &\text{otherwise}
	\end{array}\right.
\end{equation}

With the RL planning task for the CORB-Planner fully defined, 
the planning process is described by {Alg. \ref{alg}}. 
\begin{algorithm}[h]
	\caption{CORB-Planner B-spline generation}
	\label{alg}
	\begin{algorithmic}[0.8] 
		\Require grid map, vehicle motion 
		\State Initialize $p_0,p_1,p_2$ by vehicle motion and \ref{initialpoints}
		\State Initialize reference polyline $u^*$ with A* search and \ref{alg1}  
		\State Initialize SFC with $u^*$, the grid map and \ref{globalsfc},\ref{sfcg2}
		\State Composing $s_2$ with $A^*$, SFC,  $p_0,p_1,p_2$ and time
		\For {$t=2$ {to} $T$}
		\State RL policy receives $s_t$ and output $\alpha_t$ 
		\State Get B-spline control point $p_{t+1}$ with \ref{controlpoints}, \ref{actionbound}, \ref{speedbound}
		\State Compute reward $r_t$ and termination symbol $D_t$
		\If{$D_t = 1$}
			\State Terminate B-spline generation process
		\EndIf
		\State Update $s_t$ to $s_{t+1}$ according to $p_{t+1}$
		\EndFor
		\State Composing long-term B-spline with $p_0\sim p_{T+1}$
	\end{algorithmic}
\end{algorithm}

\section{Stable and Efficient RL Training via Soft Decomposed-Critic Q}\label{SDCQ} 
The robust performance of CORB-Planner is supported by our RL training process using the SDCQ algorithm. Based on a value-driven actor-critic structure, SDCQ provides stable agents with comparable performance and inference speed to PPO while achieving training efficiency twice as fast as SAC. Additionally, a simulation environment with progressively increasing obstacle density improves the planner's robustness. In this environment, we employ an exploitation-decoupled, multiple-sampling approach to generate extra data without disrupting policy exploitation, further accelerating training and mitigating overfitting.

\subsection{SDCQ for Trajectory Planning} 
SDCQ is a value-based discrete-to-continuous RL algorithm with soft RL strategies, where a discrete Q-network replaces the actor network in the actor-critic framework. The discrete Q-network updates by aligning with the continuous Q-values of the critic. To prevent the curse of dimensionality in action discretization, the SDCQ's discrete Q-network employs a decomposed structure, discretizing each action dimension independently into multiple discrete actions. In the CORB-Planner, each action dimension is discretized into M discrete actions. The discrete action $\alpha^d_t := [\alpha^d_x \ \alpha^d_y \ \alpha^d_z]^\top$, where $\alpha^d_t \in [1, M]^3$, is mapped to continuous action $\alpha_t$ as $\alpha_t = (2\alpha^d_t + 1)/M - 1$.

The decomposed Q-network, parameterized by $\theta_d$, takes observation $s_t \in \mathcal{S}$ as input and outputs $3M$ Q-values, corresponding to the $M$ discrete actions for each dimension ($x$, $y$, $z$). Let $\alpha^d_{x,k}, \alpha^d_{y,k}, \alpha^d_{z,k} = k$ for $k \in [1, M]$ represent the discrete actions in each dimension, with $Q_d(s_t, \alpha^d_{x,k}; \theta_d)$ denoting the Q-values for the $x$-axis, and similarly for $y$ and $z$. Two policies can be used to select actions from $Q_d$: the greedy policy $\pi_G$, which maximizes expected returns, and the Boltzmann policy $\pi_B$, which maximizes entropy. During training, $\pi_B$ is employed for exploration, while $\pi_G$ is used for stable performance during trajectory execution. For the $x$-axis, the policies are defined as follows
\begin{align}
	&\pi_G(\alpha^d_{x, k}|s_t; \theta_{d}) = \left\{ \begin{array}{ll} 
		\!\!\!1,\!\!\! &\mathop{\arg \max}\limits_{i\in [1, M]}Q_d(s_t, \alpha^d_{x, i}; \theta_d) \!=\! k \\
		\!\!\!0,\!\!\!&\text{otherwise}. \\
	\end{array}\right. \\
	&\pi_B(\alpha^d_{x, k}|s_t; \theta_{d}) = \frac{\exp(\kappa Q_d(s_t, \alpha^d_{x, k}; \theta_d))}{\sum_{i=1}^{M}\exp(\kappa Q_d(s_t, \alpha^d_{x, i}; \theta_d))}&
\end{align}
where 
$\kappa$ represents the adaptive temperature that controls exploration, similar to its role in SAC. The Boltzmann policy, denoted as $\pi_B$, generates actions for use in the loss function of both the discrete Q-network ($Q_d$) and the continuous critic network ($Q_c$), with the latter parameterized by $\theta_Q$. When updating $Q_d$, the corresponding continuous Q-value from $Q_c$ is matched with each discrete Q-value, and the squared error between these values is minimized. Since $Q_c(s_t, \alpha_t; \theta_Q)$ takes the state $s_t$ and the 3D continuous action $\alpha_t$ as inputs, $\alpha^d_t$ is first sampled from $\pi_B$ and mapped to $\alpha_t$; then a single dimension in discrete action $\alpha^d_t$, such as the action of dimension $x$ is replaced by $\alpha^d_{x, k}$. The loss function $\Omega^d_{x,k}(\theta_d)$ for the action $\alpha^d_{x,k}$ is expressed as
 \begin{align} 
 \Omega^d_{x, k}(\theta_d)\! &=\! \Big[ Q_d(s_t, \alpha^d_{x, k}; \theta_d) - Q_c\big(s_t, \widetilde{\alpha}_t; \theta_Q\big) \Big]^2 
 \end{align} 
  where $ \widetilde{\alpha}_t = [(2\alpha^d_{x,k}+1)/M - 1, \alpha_y, \alpha_z] $ with $  a_y, a_z \sim \pi_B$.
  
Similarly, loss functions $\Omega^d_{y,k}(\theta_d)$ and $\Omega^d_{z,k}(\theta_d)$ are defined for the $y$ and $z$ dimensions, respectively. These individual losses for all $3M $ discrete actions are combined into a single loss function, $\Omega^d(\theta_d)$, which is optimized jointly to improve training efficiency
\begin{equation}
\Omega^d(\theta_d) = \sum_{k=1}^{M} \left[ \Omega^d_{x,k}(\theta_d) + \Omega^d_{y,k}(\theta_d) + \Omega^d_{z,k}(\theta_d) \right] 
\end{equation}

The critic network $Q_c$ is optimized using a temporal-difference loss (TD) $\Omega^Q(\theta_Q)$, which includes entropy terms. A target critic network with parameters $\theta_Q'$ is used to decouple the current and target Q-values. The TD loss is defined as
\begin{align}
\Omega^Q (\theta_Q)&= \Big(Q_c(s_t, \alpha_t; \theta_{Q}) - r_t \notag\\
  &-\gamma \big[Q_c(s_{t+1}, \alpha_{t\!+\!1}; \theta_{Q}') + \kappa\mathcal{H}_{t+1} )\big]\Big)^2
\end{align}
where $\alpha^d_{t+1} \sim \pi_B$ and $\alpha_{t+1} = (2\alpha^d_{t+1} + 1)/M - 1$. The entropy term $\mathcal{H}_{t+1}$ is given by
\begin{equation}
\mathcal{H}_{t+1} = -\sum \pi_B(\alpha^d_{t+1}|s_t; \theta_d) \log \pi_B(\alpha^d{t+1}|s_t; \theta_d) .
\end{equation}

The discount factor $\gamma$ in the TD loss controls the time horizon of updates. Both loss functions, $\Omega^d(\theta_d)$ and $\Omega^Q(\theta_Q)$, are minimized at each training step. For further details about SDCQ, including adaptive temperature tuning and additional strategies like multi-step TD and normalized importance sampling, see \cite{zhang2023soft}. Notice that SDCQ supports high discretization accuracy, such as $M\geq200$, but higher M leads to lower training efficiency. In our experiments we set $M=60$ to balance training efficiency and action precision. Section \ref{evaluation} provides an empirical validation of our algorithm and ablation studies for the choose of discretization accuracy $M$.

\subsection{Easy-to-Hard Training Environment}
 \begin{figure*}[t]
	\centering
	\includegraphics[width=7in]{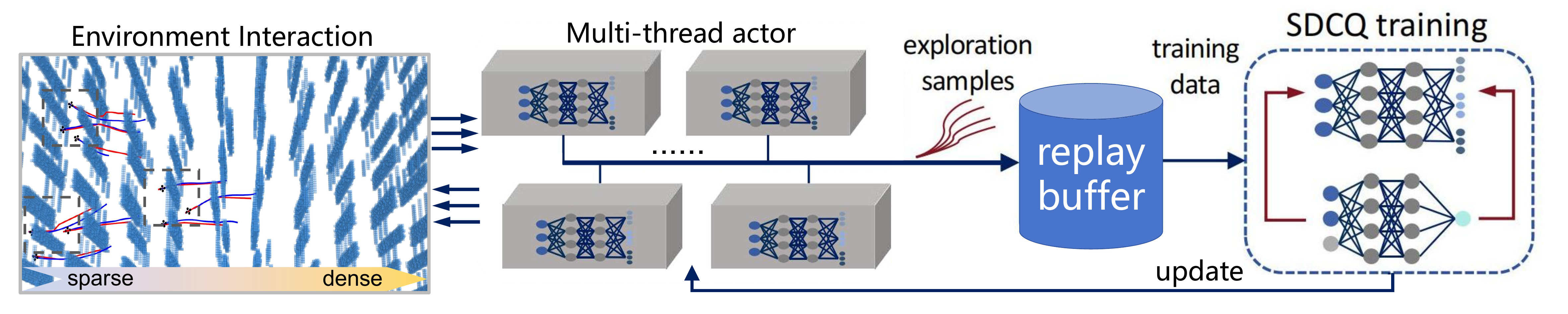}
	\caption{Training system of CORB-Planner, enables multi-thread sampling and experience replay. With only $9$ threads of simultaneous sampling, one can obtain a robust RL planner in $10$ minutes.} 
	\label{training}
\end{figure*}

The development of a well-structured simulation environment is crucial for the efficient collection of high-quality experience samples, such as dense-obstacle avoidance trajectories, while minimizing the collection of ineffective samples, such as straight-line trajectories. To achieve this objective, we have designed an easy-to-hard training environment that features obstacles with progressively increasing density, as illustrated in Fig. \ref{training}. Within this simulation training environment, random walls are configured with both horizontal and vertical features, and random gaps are introduced to allow the vehicle to pass through. The density of the walls increases progressively from left to right. Initially, the larger spacing between the random walls enhances the success rate of planning, thereby improving the quality of exploration samples. As the agent generates basic behaviors, it is progressively guided to navigate the vehicle through denser regions, thereby enhancing policy performance in more complex environments.

It is important to highlight that our simulation system does not incorporate physical simulations. The tracking of B-spline trajectories is assumed to be perfectly precise. This design choice renders the system highly deployable and capable of running on a variety of devices, including embedded platforms. To simulate real-world errors, we introduce random noise into the planning process of the CORB-Planner, rather than modifying the simulation environment itself.

\subsection{Exploitation-Decoupled Multi-Thread Sampling}
To enable effective navigation in narrow spaces, the CORB-Planner must sample experiences from challenging regions on the right side of the environment. However, biased exploration through the stochastic policy $\pi_B$ may lead to early termination due to collisions before the agent reaches these regions. To mitigate this, we propose an exploitation-decoupled experience sampling strategy. In each simulation cycle, the agent first generates an executable trajectory using the greedy policy $\pi_G$, which is free from exploration bias, and executes this trajectory via the flight controller. Concurrently, a series of non-executed exploratory trajectories is generated using $\pi_B$ and stored in the replay buffer. In each planning cycle, a $0.3$ second window is allocated to generate these non-executed trajectories, enriching the buffer with $10$ to $30$ samples. This approach improves both the quality and quantity of experiences, accelerating training and preventing overfitting.

\begin{algorithm}[H]
	\caption{CORB-Planner training process with SDCQ}
	\label{alg2}
	\begin{algorithmic}[1] 
		\Require network parameter $\theta_d, \theta_Q, \theta_Q'$, buffer $\mathcal{B}$
		\State Initialize simulation environment 
		\For {$t=0$ {to} $T$}
		\State Generate executable trajectory $B \sim \pi_G$
		\State Export $B$ to simulation flight controller
		\State Sample exploration trajectory $B_1,...,B_n \sim \pi_B$
		\State save  $B_1,...,B_n \rightarrow \mathcal{B}$
		\If{crashed}
		reset simulation environment 
		\EndIf
		\State sample a minibatch of experiences from $\mathcal{B}$
		\State Update $\theta_d, \theta_Q$ by minimizing $\Omega^d(\theta_d), \Omega^Q(\theta_Q)$
		\State Update parameters $\theta_Q'$ of the target network .
		
		\EndFor
	\end{algorithmic}
\end{algorithm}
The training of the CORB-Planner is summarized in Alg. \ref{alg2}. To further enhance training efficiency, particularly on high-performance servers, we have implemented a multi-threading mechanism to accelerate the sampling speed. Our system supports more than $10$ drones sampling trajectories simultaneously, with each drone being controlled by an independent agent. The experiences sampled by these agents are transmitted to a universal buffer, which supports centralized training and unique updates for each individual agent.
\section{Experimental Results}\label{evaluation}

CORB-Planner aims to provide effective high-speed navigation for arbitrary aerial vehicles that meet specific dynamic requirements, and show generality in various environments and tasks. This paper provides both simulation and physical results to prove the efficiency and versatility of the CORB-Planner.  

\subsection{Implementation Details}\label{implementation}
The simulation and training system for the CORB-Planner was deployed on a server equipped with an Intel 12700KF processor and an NVIDIA 4070Ti GPU. We constructed the easy-to-hard training scene with ROS Noetic. The neural networks and training code of the CORB-Planner is based on PyTorch; the policy and critic networks own two hidden layers with $256$ neurons per layer. These networks are optimized by Adam optimizer, with a learning rate of $3\times10^{-4}$. To balance exploration and exploitation, we set the target entropy of our SDCQ algorithm to $0$.
 
During training, the parameters of the reward function were set to $k_p = -30$, $k_f = 5$, and $k_s = 50$. Multiple sets of agents were trained to accommodate different maximum velocity settings. In our experiments, we selected agents with maximum velocities of $v_{max} \in\{ 4,  5,  7,  10,  15\}$. The maximum acceleration and jerk were determined based on the maximum velocity, where $a_{max} = 2 v_{max}$, $j_{max} = 50 + 10 v_{max}$. The network parameters for these agents are general for arbitrary simulation experiments and physical flights. All of the agents are trained in an unique easy-to-hard dense wall training environment, with a maximum obstacle spacing of $4.5$ m on the left and a minimum spacing of $2.75$m on the right. 

\begin{figure*}[t]
	\centering
	\includegraphics[width=7in]{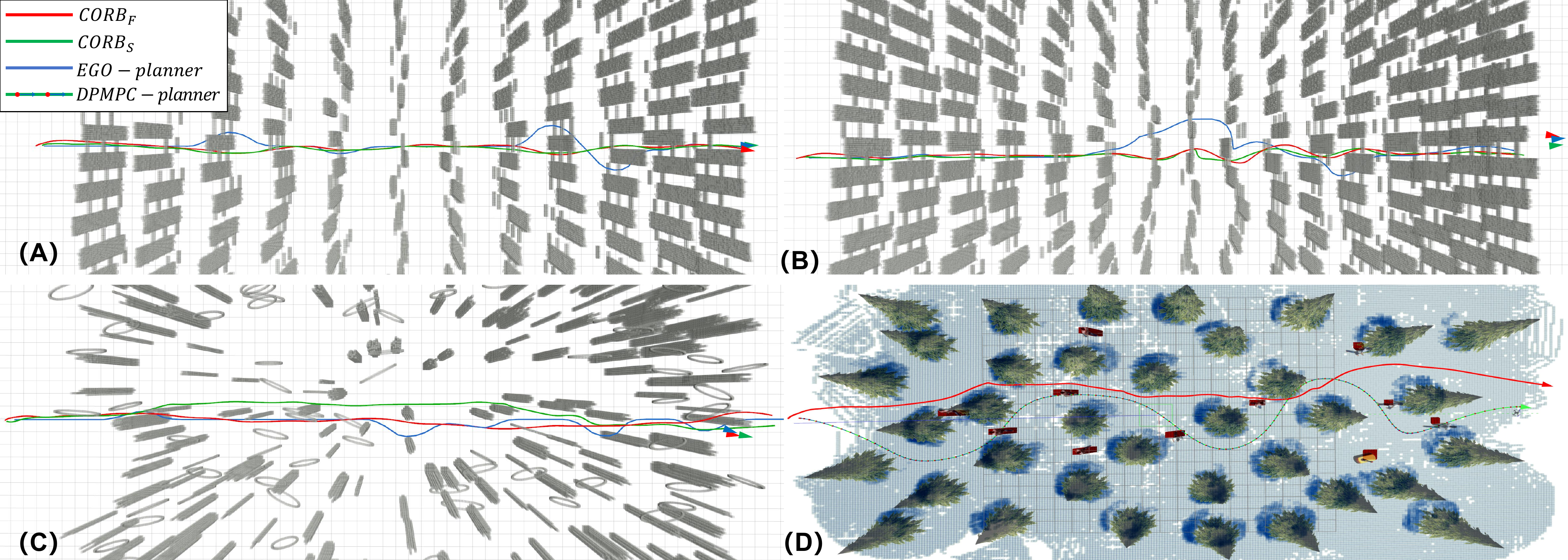}
	\caption{Behaviors of both CORB-Plnnaer and EGO-planner in the random simulation environments with maximum velocity $10$m/s: {(A)} Sparse walls; {(B)} Dense walls; {(C)} forest scene.  {(D)} Behaviors of both CORB-Plnnaer and DPMPC-Planner in the dynamic forest with walking pedestraians.} 
	\label{plannereval2}
\end{figure*}

We have prepared multiple platforms to evaluate the performance of the CORB-Planner in real-world environments, as shown at the top of Fig. \ref{fig1}. The high-performance quadrotor, tiny quadrotor, and hexarotor utilize the Livox MID360 sensor to achieve localization and mapping. An Intel N100 quad-core processor is integrated into the high-performance quadrotor and the hexarotor, while the tiny quadrotor is equipped with an ARM-based board featuring an RK3588 processor. The high-precision Point-LIO is deployed on the N100 for precise localization, and Fast-LIO2 is deployed on the ARM-based board. For the vision-based quadrotor equipped with the RealSense D430, an AMD R7-6800U octa-core processor is used to support the computational requirements of VIO. VINS-Fusion  \cite{qin2019general} is employed as the localization method for the quadrotor, and the grid map is generated based on depth images. The resolution of the grid map is set to $0.15$m across all platforms. A PID trajectory tracker, in conjunction with the PX4 autopilot, is used to provide accurate trajectory tracking.

\begin{figure*}[h]
	\centering
	\includegraphics[width=7in, height=4.5in]{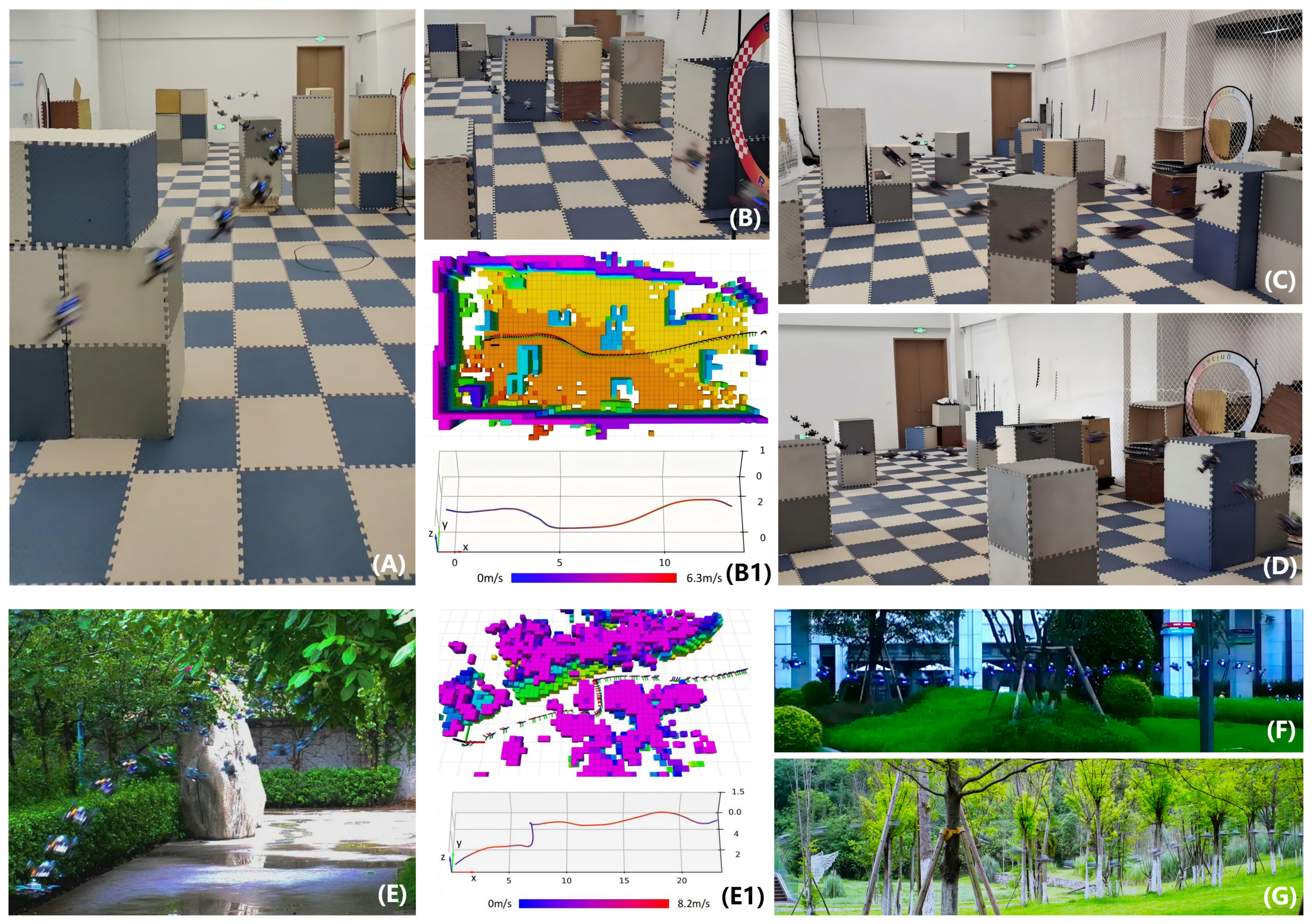}
	\caption{Physical experiments of CORB-Planner. {Indoor Experiments:} {(A)} High-performance quadrotor with the $v_{max}=10$ CORB-Planner agent; {(B)} Tiny quadrotor with $v_{max}=7$ CORB-Planner agent; {(B1)} Point cloud map and the flight trajectory in B; {(C)} Vision-based quadrotor with $v_{max}=4$ CORB-Planner agent;  {(D)} Hexarotor with the $v_{max}=7$ CORB-Planner agent.	
		{Experiments in the wlid:} {(E)} CORB-Planner navigating the high-performance quadrotor to pass through dense branches; {(E1)} Point cloud map and the flight trajectory in E; {(F)} Experiment in urban scenes; {(G)} Experiment in forest scenes.} 
	\label{generalflight}
\end{figure*}
\subsection{Benchmarks in Simulation}
\begin{table*}[t]
	\centering
	\setlength{\tabcolsep}{2.9pt}
	\renewcommand\arraystretch{1}
	\caption{Success rate and episode time cost of CORB-Planner and EGO-planner}
	\begin{tabular}{cccccccccc} 
		\hline
		{algorithm} &\small{CORB$_F$} &\small{CORB$_S$} &\small{EGO-planner} &\small{CORB$_F$} &\small{CORB$_S$} &\small{EGO-planner} &\small{CORB$_F$} &\small{CORB$_S$} &\small{EGO-planner}\\
		
		{$v_{max}$} &\small{$15$m/s} &\small{$15$m/s} &\small{$15$m/s} &\small{$10$m/s} &\small{$10$m/s} &\small{$10$m/s} &\small{$7$m/s} &\small{$7$m/s} &\small{$7$m/s}\\
		\hline

		\small{forest scene} & \small{17/20, \textbf{6.7s}} & \small{19/20, 9.3s} & \small{19/20,11.4s} &  \small{19/20, \textbf{8.2s}} & \small{19/20, 10.2s} & \small{\textbf{20/20},11.4s} &  \small{20/20,13.2s} & \small{20/20, \textbf{13.1s}} & \small{20/20, 14.1s} \\
		
		\small{sparse walls} & \small{18/20, \textbf{7.5s}} & \small{\textbf{20/20}, 7.7s} & \small{18/20, 10.4s} &  \small{17/20, \textbf{9.3s}} & \small{20/20, 9.6s} & \small{20/20, 13.2s} &  \small{20/20, \textbf{10.4s}} & \small{20/20, 12.7s} & \small{20/20, 15.6s} \\
		
		\small{dense walls} & \small{10/20, \textbf{7.9s}} & \small{\textbf{18/20}, 8.9s} & \small{3/20, 13.6s} &  \small{15/20, \textbf{9.4s}} & 
		\small{\textbf{19/20}, 11.1s} & \small{7/20, 16.3s} & \small{20/20, \textbf{13.3s}} & \small{20/20, 13.9s} & \small{13/20, 17.9s} \\
		\hline
		\label{algt}
	\end{tabular}
\end{table*}

We compared CORB‑Planner against optimization‑based EGO‑planner in two types of simulated environments: 1)
Random Forests: $200$ cylindrical and circular obstacles, as used in prior EGO‑planner evaluations; and, 2)
Easy‑to‑hard wall courses: Sparse walls with inter‐wall spacings of $5$m to $3.5$m, and dense walls with spacings of $4$m to $2$m.

For each environment, we ran $20$ episodes at maximum speeds $v_{max} \in  \{7, 10, 15\}$. Two CORB‑Planner variants were tested:
\begin{itemize}
    \item fast CORB (denoted as CORB$_F$) with collision penalty $k_p = -30$  and SFC‑follow reward $k_f = 8$, favoring aggressive trajectories.
    \item  Safe CORB (denoted as CORB$_S$) with $k_p = -50$, $k_f = 3$,  favoring conservative behavior.
\end{itemize}
All planners used $a_{max} = 2v_{max}$ and $j_{max} = 50 + 10v_{max}$. The vehicle’s task was to travel from 
$(0,-32)$ to $(0,32)$. Table I summarizes success rates and average episode times. Compared to EGO‑planner—which tends to generate conservative, collision‑averse paths—CORB$_F$ completes courses significantly faster but with a higher failure rate. CORB$_S$ matches EGO‑planner’s success rate while maintaining a clear time advantage.

We also evaluated dynamic‐obstacle avoidance in a ``moving forest” scenario, with nine pedestrians navigating at up to $v_{max}=1$m/s. CORB‑Planner (set to $4$m/s) models pedestrians in the occupancy grid; upon an infeasible A$^*$ solution, the planner commands a hover until a path reappears. In $20$ trials, both CORB‑Planner and the DPMPC‑planner \cite{xu2022dpmpc} avoided all moving obstacles, confirming CORB‑Planner’s ability to handle slowly moving obstacles at higher speeds. Fig. \ref{plannereval2} visualizes sample trajectories across all scenarios.

\subsection{Ablation Studies}

\begin{figure}[H]
	\centering
	\includegraphics[width=3.5in,height=1.9in]{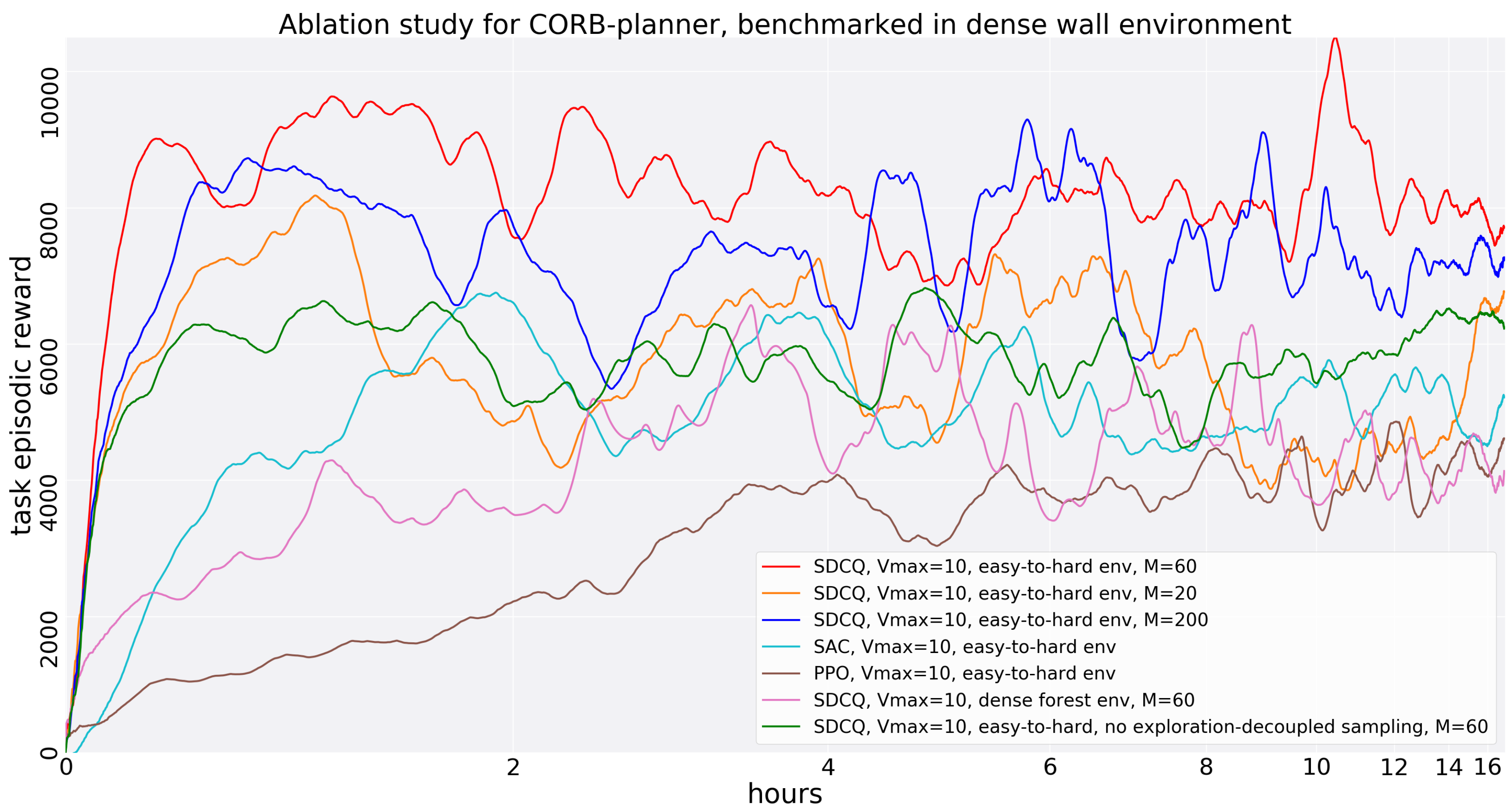}
	\caption{
		Results of ablation studies: (A) comparison of RL algorithms (PPO, SAC, SDCQ); (B) impact of training environment (random forest vs.\ easy‑to‑hard curriculum);(C) impact of exploration-decoupled sampling; (D) effect of SDCQ discretization level \(M\) on performance. 
	} 
	\label{plannereval1}
\end{figure}

To quantify the impact of our algorithmic and training design choices, we conducted four ablation experiments (all with $v_{max}=10$, evaluating every five minutes over three episodes each). Results are plotted in Fig.
\ref{plannereval1}, which each curve represents the average of five training attempts:
\begin{itemize}
    \item RL Algorithm Comparison: SDCQ vs. PPO and SAC. SDCQ converges in under $10$ mins and achieves higher final success rates than both actor‑critic baselines.
\item Training environment: Easy‑to‑hard curriculum vs. static random forest. Curriculum learning accelerates convergence and yields stronger policies in dense scenarios.
\item Exploration‑decoupled sampling: With vs. without non‑executed exploratory trajectories. Decoupled sampling improves sample diversity and final performance.
\item Discretization level 
$M$ in SDCQ: Varying 
$M$ from low to high. While SDCQ allows fine discretization, 
$M=60$ strikes the best balance between action precision and training efficiency.
\end{itemize}

\subsection{Generalization Tests on Physical Platforms}

We deployed CORB‑Planner on four distinct UAVs—ranging from LiDAR‑equipped quadrotors (Livox MID360) and vision‑based quadrotors (Realsense D435i) to hexarotors—using agents trained for $v_{max} \in\{ 4,  5,  7, 10\}$m/s.
Indoor flight tests measured each platform’s practical speed ceiling and robustness in clutter. Fig. \ref{generalflight} presents representative flight paths, demonstrating that CORB‑Planner generalizes across airframes and sensing configurations while maintaining high success rates and real‑time onboard performance.

All four drones can effectively avoid the obstacles with $v_{max} = 4$. Due to the drawback of visual-inertial odometry, the visual-based quadrotor may crash into the obstacles with the $v_{max} \geq 5$ agent. With LiDAR-based SLAM methods, the tiny quadrotor and the hexarotor can avoid the obstacles successfully with $v_{max} \in\{ 4,  5, 7\}$. The high-performance quadrotor with the highest dynamic performance can associate with all four agents. The results show that the capability of a CORB-Planner agent on a specific platform is determined by the dynamic performance and accuracy of the SLAM methods; agents with low velocity and acceleration can be widely applied to various types of vehicles.


\subsection{Autonomous Flight in Complex Environments}

The CORB-Planner is designed to achieve generalization across both diverse platforms and environmental conditions. To evaluate its performance, high-speed trials were conducted to assess the planner’s capability to navigate complex unstructured environments. Fig. \ref{generalflight}(E) illustrates a dense forest scenario where CORB-Planner, deployed on a high-performance quadrotor, generated smooth trajectories for obstacle avoidance. During this test, the quadrotor traveled $23$m in $4.2$s, reaching a peak velocity of $8.2$m/s. In narrow corridors, the planner autonomously reduced flight speed to minimize collision risks. The corresponding grid map and vehicle trajectory are shown in Fig. \ref{generalflight}(E1). Furthermore, experiments carried out in urban and forest settings, as shown in Figs. \ref{generalflight}(F) and \ref{generalflight}(G), respectively, demonstrate the adaptability and robustness of the planner to varying environmental conditions.

\subsection{Dynamic Obstacle Avoidance in Physical Environment}
\begin{figure}[th]
	\centering
	\includegraphics[width=3.4in]{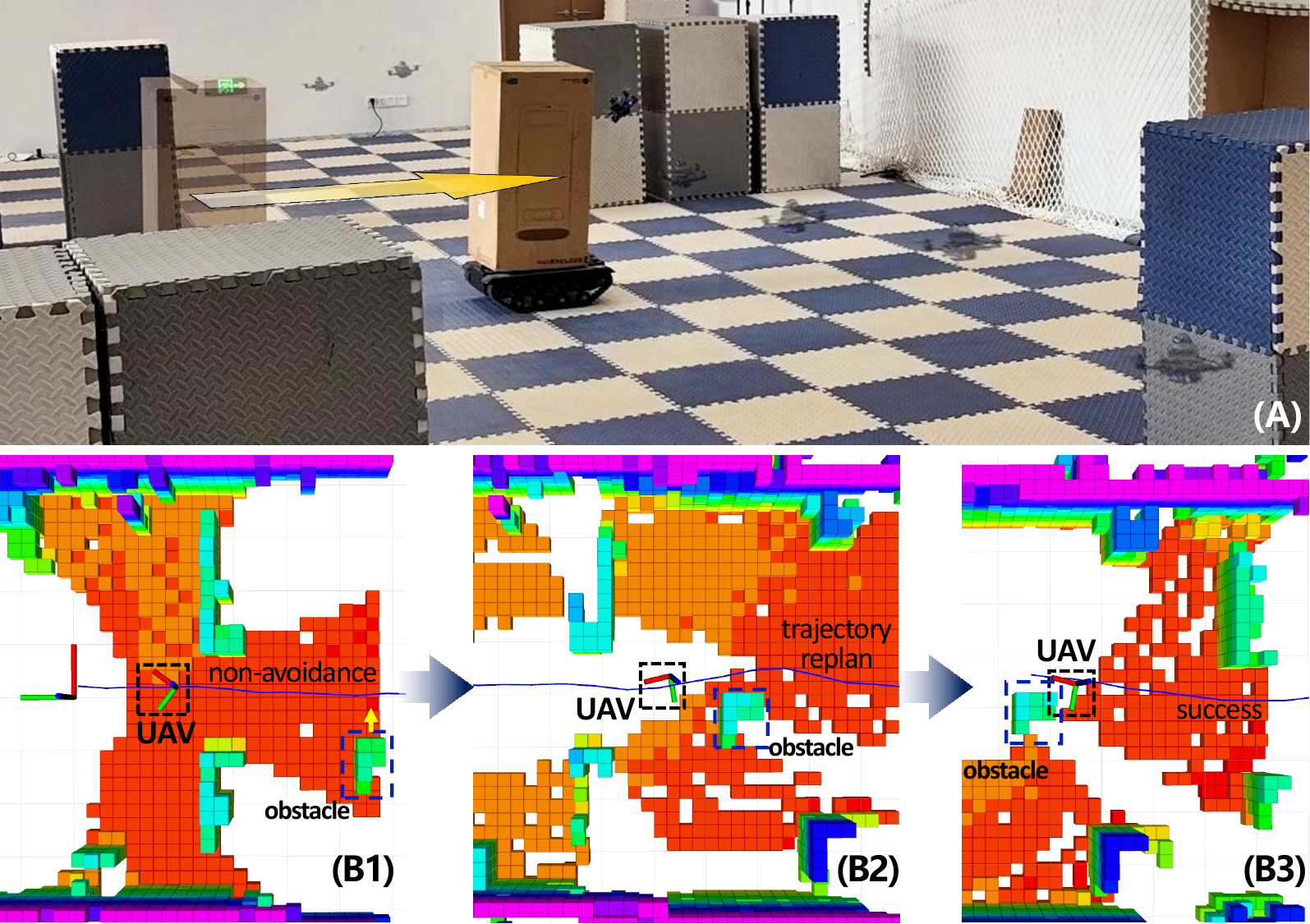}
	\caption{(A) Physical performance of moving obstacle avoidance. (B1)$\rightarrow$(B2)$\rightarrow$(B3)Process of CORB-Planner avoiding the dynamic obstacle.} 
	\label{dynamicflight}
\end{figure}

To evaluate the capability of the CORB-Planner in avoiding dynamic obstacles within physical environments, we conducted an experiment involving a ground vehicle moving at an average speed of $1$m/s, which served as a moving barrier. Additionally, our high-performance quadrotor, with a maximum velocity of $4 $m/s, was employed to execute the task. The experimental results are illustrated in Fig. \ref{dynamicflight}. When the obstacle entered the quadrotor's trajectory, the CORB-Planner generated a real-time avoidance trajectory to bypass the obstacle, as depicted in Fig. \ref{dynamicflight}(B2).

\subsection{CORB-Planner on a Lightweight Platform}

\begin{figure}[th]
	\centering
	\includegraphics[width=3.4in]{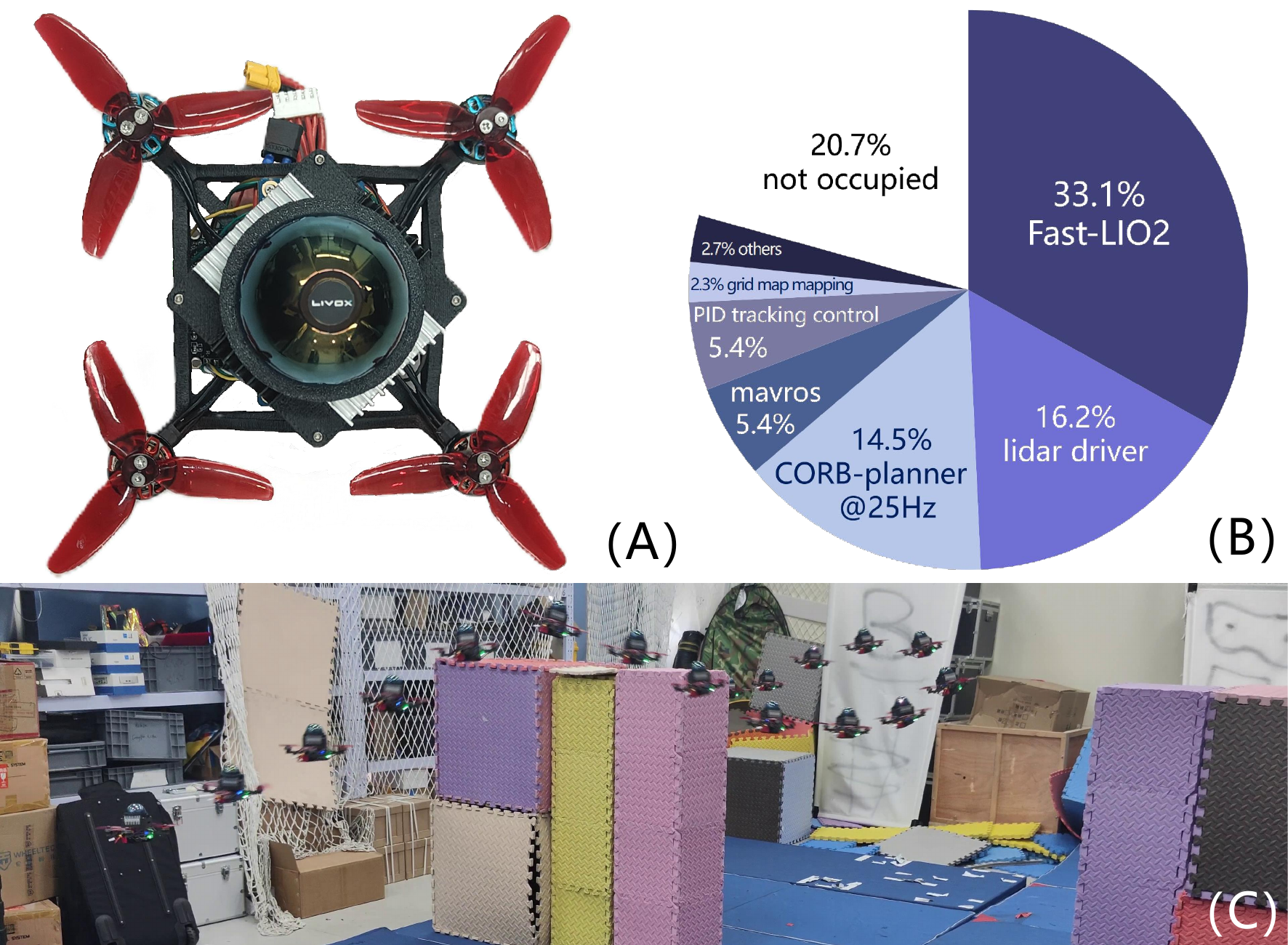}
	\caption{CORB-Planner on the ultra lightweight quadrotor. (A) physical platform; (B) CPU consumption during the flight; (C) flight trajectory} 
	\label{flight618}
\end{figure}

To validate the feasibility of deploying the CORB-Planner on lightweight platforms, we developed a lightweight computing quadrotor weighing only $275$g. The platform is equipped with a $65$mm$\times30$mm Orange Pi Zero $2$W tiny board, which features an Allwinner H618 processor with quad-core Cortex-A53. The primary sensor on the platform is a Livox MID360 LiDAR, whose weight was reduced to $132$g. A lightweight version of Fast-LIO2 was implemented on the platform to provide localization and mapping capabilities. We evaluated the performance of our lightweight computing quadrotor in an indoor environment. At a maximum velocity of $2$m/s and a replanning frequency of $25 $Hz using the CORB-Planner, the quadrotor was able to effectively avoid obstacles. During flight, Fast-LIO2 consumed $33.1$\% of the CPU resources, while the CORB-Planner required only $14.5$\%. Computational consumption of each module is illustrated in Fig. \ref{flight618}.
\

\section{Conclusions}
In this paper, we proposed a lightweight and robust RL planning method CORB-Planner for aerial vehicles, which takes flight status and SFC as input to plan smooth B-spline trajectories. We utilized the SDCQ algorithm for simulation training to obtain effective path-planning policies in several minutes, and the simulation policy can be directly applied to physical vehicles in various environments. Physical evaluations have proven that our algorithm is general to various of vehicle platforms, sensors and environments, and effectively avoid arbitrary obstacles during high-speed autonomous flight.

\bibliographystyle{IEEEtran}
\bibliography{references1.bib}

\end{document}